\documentclass{article}

\usepackage{arxiv}

\usepackage[utf8]{inputenc} 
\usepackage[T1]{fontenc}    
\usepackage{hyperref}       
\usepackage{url}            
\usepackage{booktabs}       
\usepackage{amsfonts}       
\usepackage{nicefrac}       
\usepackage{microtype}      
\usepackage{lipsum}
\usepackage{graphicx}
\usepackage{amsmath}
\usepackage{mathtools}
\usepackage{amssymb}
\usepackage{latexsym}
\usepackage[table,xcdraw]{xcolor}
\usepackage{multirow}
\usepackage[ruled,vlined]{algorithm2e}
\usepackage{comment}
\usepackage{threeparttable}

\usepackage{url}
\usepackage{xcolor}
\graphicspath{ {./images/} }

\title{Reconfigurable Cyber-Physical System for Critical Infrastructure Protection in Smart Cities via Smart Video-Surveillance}

\author{
 Juan Isern \\
  Computer Architecture and Technology CITIC\\
  University of Granada\\
  Granada 18014, Spain \\
  \texttt{jisern@ugr.es} \\
  \And
 Francisco Barranco \\
  Computer Architecture and Technology CITIC\\
  University of Granada\\
  Granada 18014, Spain \\
  \texttt{fbarranco@ugr.es} \\
  \And
 Daniel Deniz \\
  Computer Architecture and Technology CITIC\\
  University of Granada\\
  Granada 18014, Spain \\
  \texttt{danideniz@ugr.es} \\
  \And
 Juho Lesonen \\
  Visidon Ltd. \\
  University of Granada\\
  Oulu 90590, Finland \\
  \texttt{juho.lesonen@visidon.fi} \\
  \And
 Jari Hannuksela \\
  Visidon Ltd. \\
  University of Granada\\
  Oulu 90590, Finland \\
  \texttt{jari.hannuksela@visidon.fi} \\
  \And
 Richard R. Carrillo \\
  Computer Architecture and Technology CITIC\\
  University of Granada\\
  Granada 18014, Spain \\
  \texttt{rcarrillo@ugr.es} \\
  
}

\begin{document}
\maketitle
\begin{abstract}
Automated surveillance is essential for the protection of Critical Infrastructures (CIs) in future Smart Cities. The dynamic environments and bandwidth requirements demand systems that adapt themselves to react when events of interest occur. We present a reconfigurable Cyber Physical System for the protection of CIs using distributed cloud-edge smart video surveillance. Our local edge nodes perform people detection via Deep Learning. Processing is embedded in high performance SoCs (System-on-Chip) achieving real-time performance ($\approx$ 100 fps - frames per second) which enables efficiently managing video streams of more cameras source at lower frame rate. Cloud server gathers results from nodes to carry out biometric facial identification, tracking, and perimeter monitoring. A Quality and Resource Management module monitors data bandwidth and triggers reconfiguration adapting the transmitted video resolution. This also enables a flexible use of the network by multiple cameras while maintaining the accuracy of biometric identification. A real-world example shows a reduction of $\approx$ 75\% bandwidth use with respect to the no-reconfiguration scenario.
\end{abstract}


\section{Introduction}
\label{sec:int}

Smart Cities use intelligent computing technologies to gather and analyze data from multiple sources to optimize resources, monitor activities, or prevent potential risks while improving the services they provide to its citizens \cite{washburn2009helping}. Critical Infrastructures are vital resources for the development of a society, their failure would have a very high impact/cost. \cite{Gupta2014}. These include roads, communications, water, energy, etc. According to \cite{Talari2017}, security is \textit{the most relevant element of the Smart Cities from the citizen's point of view}. Even a minor disruption, either accidental or deliberate, will degrade the system performance and inflict significant economic and social losses.

Smart automated surveillance plays a crucial role in Critical Infrastructure Protection (CIP) and represents a major challenge for the next years. Modern approaches include Cyber-Physical Systems (CPS) that integrate intelligent computing devices and networks that constantly monitor the infrastructures to guarantee the adequate level of security \cite{Cenedese2014}.

After the launch of AlexNet \cite{Krizhevsky2012}, Convolutional Neural Networks (CNN) and other Deep Learning (DL) techniques are increasingly used in many areas of research, often improving the results present in the state of the art until their appearance. Thus, we find application of DL techniques in fields such as: Natural language processing (NLP) (e.g. sentence modeling \cite{Kalchbrenner2014}), financial forecasting (e.g. Analysis of gramian angular fields (GAF) images from the Standard \& Poor's 500 index \cite{Barra2020}), point cloud analysis (e.g. Relation-Shape CNN \cite{Liu2019}), computer-aided medical diagnosis (e.g. brain tumor classification with magnetic resonance imaging (MRI) \cite{Abiwinanda2019}), etc. Biometrics, and specifically face recognition plays a crucial role in CPS providing a way to automate access control. The neural network based algorithms are much more accurate (for the Labeled Faces in the Wild - LFW dataset \cite{LFW2007}, \textit{LFW} accuracy $>99\%$) than previous methods such as eigen vector \cite{Turk1991} (\textit{LFW} accuracy $\approx60\%$), and Local Binary Pattern (LBP) \cite{Ahonen2006} based algorithms (\textit{LFW} accuracy $\approx70\%$). Another of the uses that the application of neural networks has in this field is the clustering of facial attributes extracted by CNN \cite{Abate2020, liu2015faceattributes}. This technique would allow to obtain an identikit of each detected subject with an intermediate quality of the image, so it could improve their facial re-identification.

Classic video surveillance using closed circuit television (CCTV) systems have become an essential element for security and law enforcement. However, classic CCTV surveillance systems depend on the attention capacity of a human operator, who is confronted with a collage of many video streams \cite{Haering2008}. Thus, automatic video analysis is required to appropriately scale to the CIs of a Smart City. Otherwise, massive CCTV becomes only a forensic resource to identify the cause after an accident or failure.

Some limitations in conventional video surveillance industry include: 1) High cost associated to the need of personnel watching the videos, currently being replaced by smart automatic systems. Also, fatigue-related errors are a significant issue. Machine Learning (ML) is the dominant trend for automatic video analytics but it requires large datasets for training, not available for this field. 2) Multi-camera systems produce a huge amount of data demanding high bandwidths, thus local/distributed processing is required since centralized processing cannot scale. 3) High latency, communication delays are a major concern in latency-sensitive surveillance, real-time data processing guarantees instant decision making to reduce risks and facilitate pro-active security \cite{Chen2018}. 4) High cost of the deployment of ad-hoc communication networks with high data bandwidth and the lack of flexibility after deployment of rigid centralized systems.

Contrarily to centralized fully human-operated systems, the demand for distributed smart video surveillance systems increases. Recently, the fast evolution of high performance devices has opened the possibility of video analytics embedded in CPS edge devices. These new local processing capabilities foster building new distributed paradigms for CPS and tackle problems such as network traffic congestion, high latencies, and helps reducing costs of staff and network infrastructures for high data bandwidth \cite{Chen2018}. Local edge nodes enable embedding processing of complex tasks that are now done locally, reducing the communication with central nodes and thus the communication traffic. Also, local processing partially reduces latency at least for the tasks performed locally. Central nodes are in charge of gathering results from local nodes, performing analysis, and making decisions about the overall system performance. Beyond that, CPS are also expected to be context-aware and dynamically reconfigure themselves adapting to the world in real time \cite{Masin2017}.

The use of multi-camera systems increases the complexity of the system, forcing it to enable mechanisms to characterize the 3D multi-view geometry for perimeter control, complement the biometric identification via facial recognition with additional cues, predict trajectories or perform robust tracking of moving targets across multiple views \cite{Salman2018}. The central cloud node facilitates the communication between the nodes and accordingly allows for the reconfiguration of the local nodes for carrying out these high-level tasks.

The main contributions of our smart video-surveillance CPS are: 1) a multi-view system, scalable and easily calibrated which automatically triggers alarms for potential risks; 2) the integration of different ML and computer vision techniques in the same video surveillance workflow, trained and tested with a wide variety of video surveillance scenes; 3) the reduction of processing and network bandwidth, using a distributed CPS with local edge nodes that are dynamically reconfigured according to the task conditions or metrics; 4) real-time video processing using high-performance embedded devices such as the Nvidia Jetson TX2 and Jetson Xavier Systems-on-Chip (SoCs).

\begin{figure}[t] 
    \centering
    \includegraphics[scale=.47]{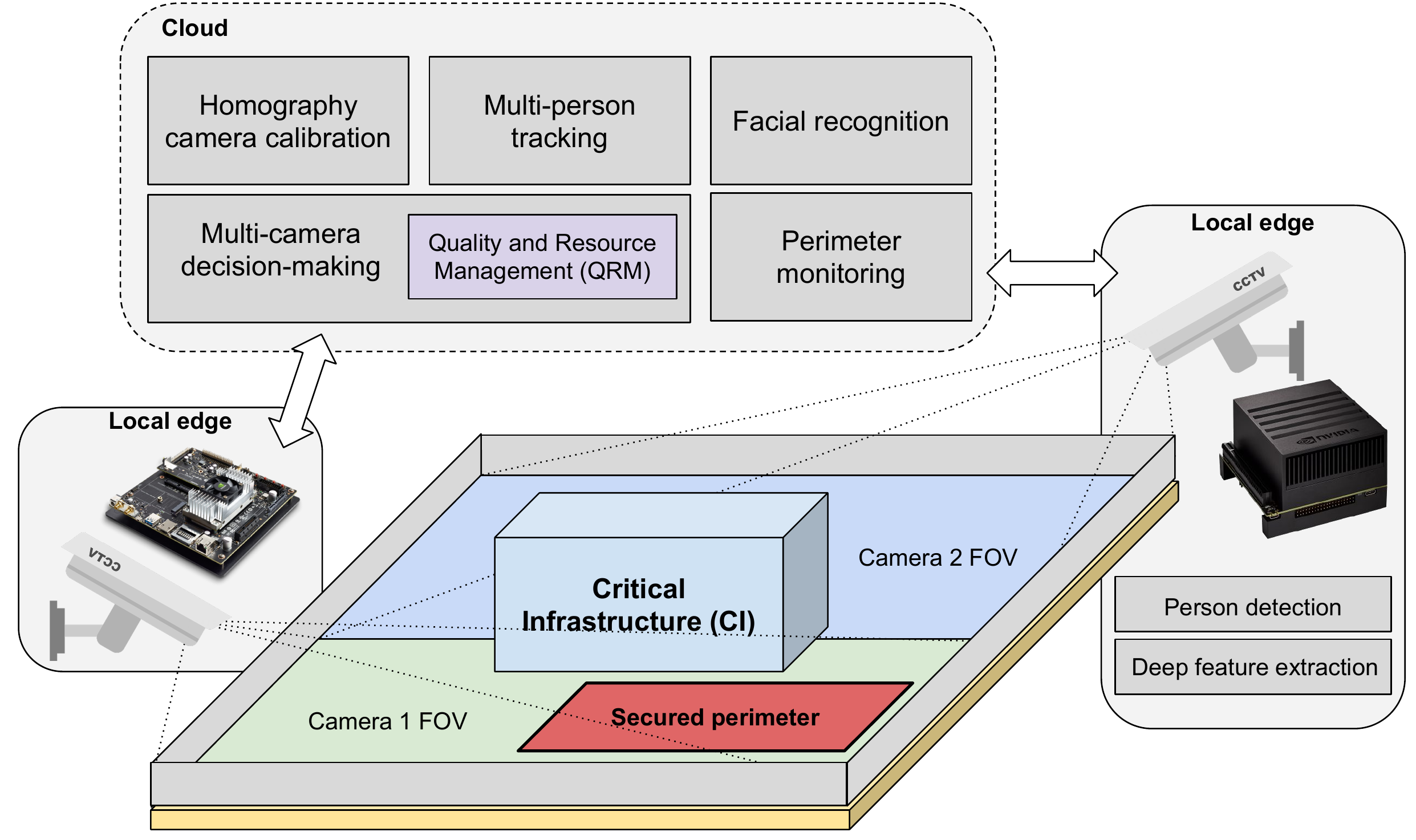}
    \caption{Overview of our reconfigurable CPS for CIP in a heterogeneous cloud-edge architecture: distributed local edge nodes process video embedded in high-performance SoCs (left: Nvidia Jetson TX2, right: Nvidia Jetson Xavier), performing human detection and extraction of robust features via Deep Learning (DL); the cloud server carries out perimeter monitoring, biometric facial recognition and tracking, using robust features, multi-camera management using homographic information, and makes decisions about the edge optimal configuration based on the Quality and Resource Management information.}
\label{fig:overview}
\end{figure}

\section{Proposed Approach}
\label{sec:apr}
In this paper we propose a heterogeneous reconfigurable CPS with distributed computing between a central server and networked SoC nodes (see Fig. \ref{fig:overview}). Each node is connected to a camera from which it obtains livestream video of a monitored area,within the critical infrastructure. This video is pre-processed locally by the node to detect people within the image and obtain a robust descriptor of their appearance, applying DL techniques. Next, the people's locations and their descriptions , together with the video, are sent to the cloud from each node. In the cloud, the calibrated system of camera nodes let us estimate the position in the real world by homography.  Location and appearance cues are fed into our tracking component  that is complemented by a component that also monitors the secure perimeter of the infrastructure. In addition, facial recognition of the people detected is performed in order to classify operators/intruders. The cloud also makes decisions about quality and resource management and it is responsible for triggering the reconfigurations of the nodes, according to existing surveillance needs. In our example in this work, given the massive amount of data transmitted by multiple cameras, reconfiguration is triggered by data bandwidth usage.

\begin{figure}[t]
\centering
\includegraphics[scale=.42]{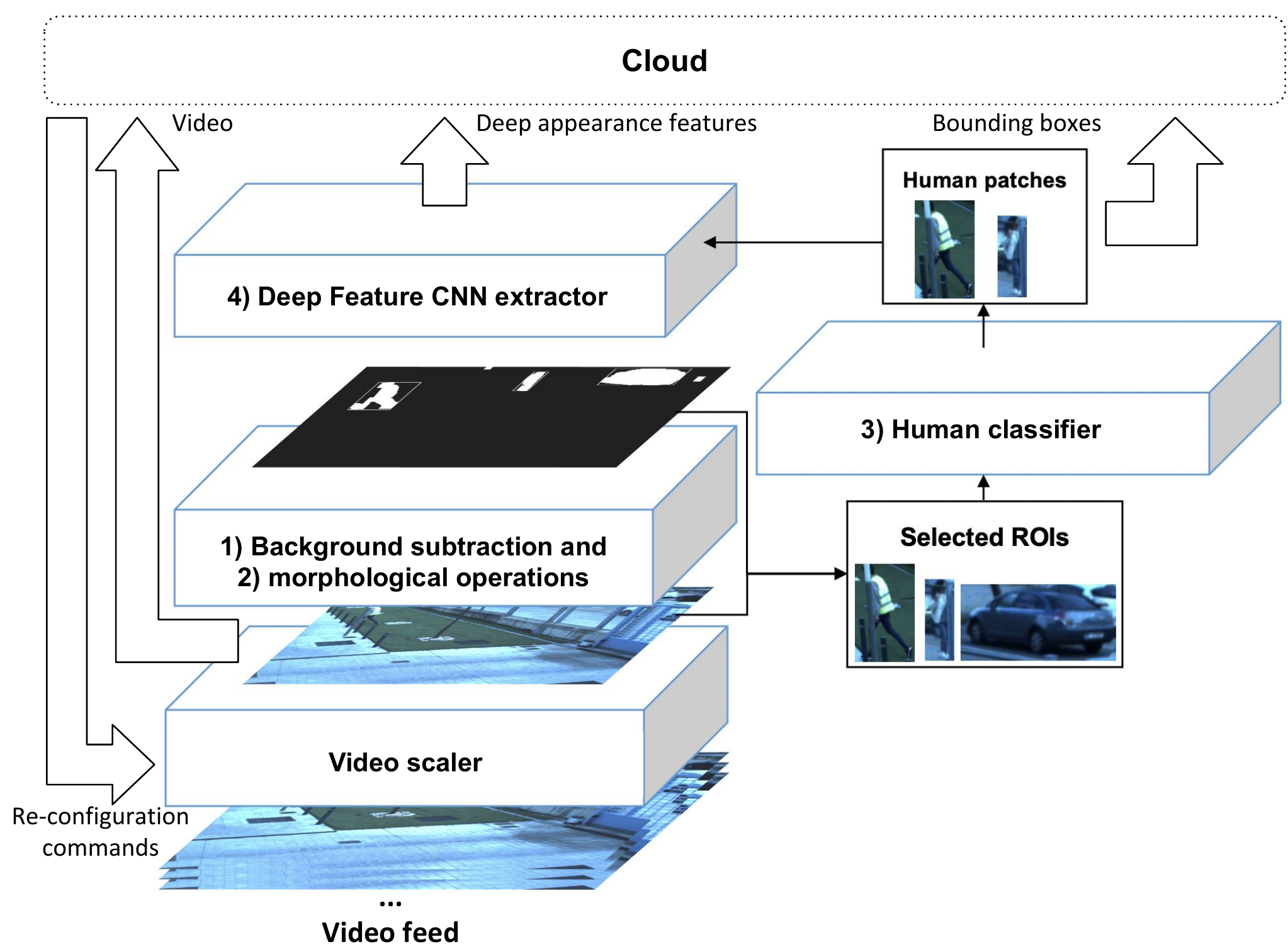}
\caption{Overview of local edge live video processing: 1) Mixture of Gaussians (MOG) background subtraction determines ROIs; 2) smaller regions that correspond to noise are removed by morphological operations; 3) each ROI is fed into the CNN human detector (in the example, ROIs include two operators of the CI and a car); 4) ROIs classified as people are passed to a deep feature extractor for target disambiguation. Finally, the video, bounding boxes containing positive examples (people), and deep feature descriptors are sent to the cloud for further processing.}
\label{fig:edge}
\end{figure}

\section{Material and methods}
\label{sec:mat}
The continuous advances in high-performance computing and embedded devices becoming lighter, more portable, and interconnection capabilities boost the development of CPS \cite{Scionti2019}. Modern CPSs are becoming distributed architectures of highly connected intelligent embedded systems that are getting increasingly autonomous. Moreover, CPS are dynamic systems which need to adapt to changing external conditions or monitored qualities triggering reconfigurations in real time \cite{Masin2017}.

Smart surveillance has become a popular CPS application, due to the real-time processing requirements and the large number of cameras needed to cover a large FOV (field of view). Processing-intensive tasks and its real-time requirements motivate a distributed edge-cloud architecture with optimized edge nodes (high performance SoCs) integrating efficient real-time processing, and the cloud that manages edge devices and performs task offloading \cite{Scionti2019}. 

The following subsections describe the components of our reconfigurable CPS. The two parts that conform the processing at the edge are depicted in Fig. \ref{fig:edge}: the human detection component marks image regions where humans are located, and then DL features are extracted from these regions. At the cloud level, DL features complement the biometric identification via facial recognition, specially useful for reidentification e.g. when a person re-entries the FOV. The target regions are used for multi-camera tracking. To ensure the consistency, calibration via homography enables the use of a common frame of reference. Finally, reconfiguration and QRM module are described in section \ref{sec:rCPS}.

\subsection{DL for human detection}
\label{subsec:dnn}

\begin{figure}[ht]
\centering
\includegraphics[scale=.75]{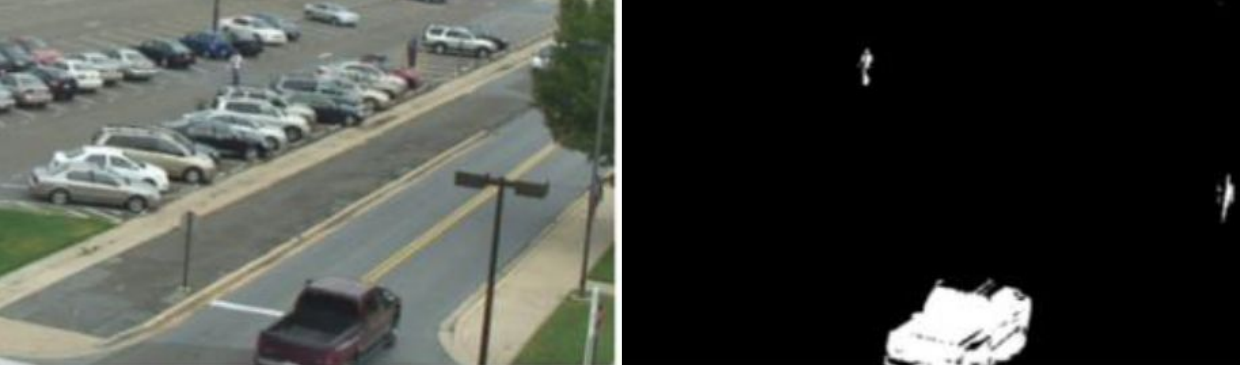}
\caption{Background substraction using MOG for traffic sequence (VIRAT dataset): moving objects (foreground) are separated from the background; the foreground in this scene pops out as the black pickup truck (bottom) and two people (parking lot and sidewalk).}
\label{fig:fgbgsubstraction}
\end{figure}

One of the common approaches for detecting and tracking moving objects in a video stream obtained by a static camera is background subtraction. It extracts a region of interest (ROI) of the object in movement within a scene, defined as foreground, generating a model of the background of the image with those objects that remain static \cite{Piccardi2004}. Fig. \ref{fig:fgbgsubstraction} illustrates the use of background subtraction applying the Mixture of Gaussians (MOG) approach \cite{Zivkovic2004}. This method models the history of each pixel value as a set of Gaussian distributions whose parameters are continuously updated with new incoming frames from a video sequence. The mathematical model that this method uses is shown in Eq. \ref{eq:mog}.

\begin{equation}
\hat{p}\left ( \vec{x} | \chi_{T}, BG+FG \right ) = \sum_{m=1}^{M} \hat{\pi}_{m} \nu \left ( \vec{x}; \hat{\vec{\mu}}_{m}; \hat{\sigma}_{m}^{2}I \right )
\label{eq:mog}
\end{equation}

For the above equation, $\hat{\vec{\mu}}_{1},...,\hat{\vec{\mu}}_{M} $ are the estimates of the mean and $ \hat{\sigma}_{1},...,\hat{\sigma}_{M} $ the estimates of the variance of the different Gaussian distributions of each pixel value. Additionally, $ \hat{\pi}_{m} $ is the weight that determines the importance of each distribution. This method runs continuously in real-time in our system at the node, estimating the potential positions of ROIs to be classified as humans or otherwise discarded.

Human detection approaches are divided into classic methods that use handcrafted features and novel DL methods that automatically learn these features from large annotated datasets. DL-based human detection approaches outperform the detection task in comparison to classic approaches such as the most common Support Vector Machine (SVM-) based classification methods \cite{Park2020}. Particularly, CNNs automatically learn high-level semantic characteristics from data through structure, exhibiting superior performance in classification tasks \cite{Zhao2019}. Although training a CNN is compute-intensive, techniques such as Transfer Learning alleviates this problem, allowing a pre-trained model to be partially retrained for different applications. In this framework, the weights of a network that is very successful differentiating low- and medium-level image features are re-used. Then, partial re-training updates the weights of the last network layers that extract high-level features to adapt the network to a different classification problem. In this work, we perform human detection via CNNs, retraining a network using the Transfer Learning technique to adapt its original weights (trained with the ImageNet dataset \cite{imagenet_cvpr09}) to our video surveillance problem.

We chose the \textit{MobileNetV2} \cite{Sandler2018} as base model, popular for embedded applications and with a good accuracy vs. resource trade-off. This model performs a convolution for each channel individually instead of all of them at once, reducing the number of calculations and thus, making it more suitable for architectures with limited resources. 

\subsection{Camera calibration with homographic transformation}
\label{subsec:homography}
Calibration estimates camera intrinsic parameters through the correspondence between 3D points in the camera image plane, and the relationship between cameras \cite{Long2019}.
Knowing this relationship is essential e.g. to estimate target trajectories out of the FOV, determine the position of targets or delimit the secured perimeter across different cameras. We use homography because, among the classical calibration methods, 2D planar target calibration methods provide high accuracy and flexibility \cite{Zhang2000}.

A homography is any projective transformation that determines a geometric correspondence between two planes. In our case, the homography provides a common frame of reference for the accurate location of targets across cameras. 
Given a set of points corresponding to each other $ x_{i}\leftrightarrow{x_{i}}' $ where $ x_{i} $ comes from the position in the camera view, while $ {x_{i}}' $ is the position corresponding to that point on the plane of the scene and by writing $ {x_{i}}' = \left (\chi_{i}', \varphi_{i}', \omega_{i}' \right )^{T} $ with homogeneous coordinates, we can estimate the homography matrix $ H $ between the view and the plane as $ {x_{i}}' \times Hx_{i} = 0 $. Thus, for each pair of corresponding points, three linear equations are written as shown in Eq. (\ref{eq:homography}), with $ {h_{i}}, i=1,2,3 $ a $ 3 \times 1 $ vector made of the elements of the i-th row of $ H $.

\begin{equation}
\begin{bmatrix}
0^{T} & -\omega_{i}'x_{i}^{T} & -\varphi{i}'x_{i}^{T} \\ 
\omega_{i}'x_{i}^{T} & 0^{T} & -\chi_{i}'x_{i}^{T} \\ 
\varphi{i}'x_{i}^{T} & \chi_{i}'x_{i}^{T}  & 0^{T}
\end{bmatrix} 
\begin{pmatrix}
h_{1} \\ 
h_{2} \\ 
h_{3}
\end{pmatrix}
= 0
\label{eq:homography}
\end{equation}

\subsection{Facial recognition}

 \begin{figure}[ht]
 	\centering
 	\includegraphics[scale=.5]{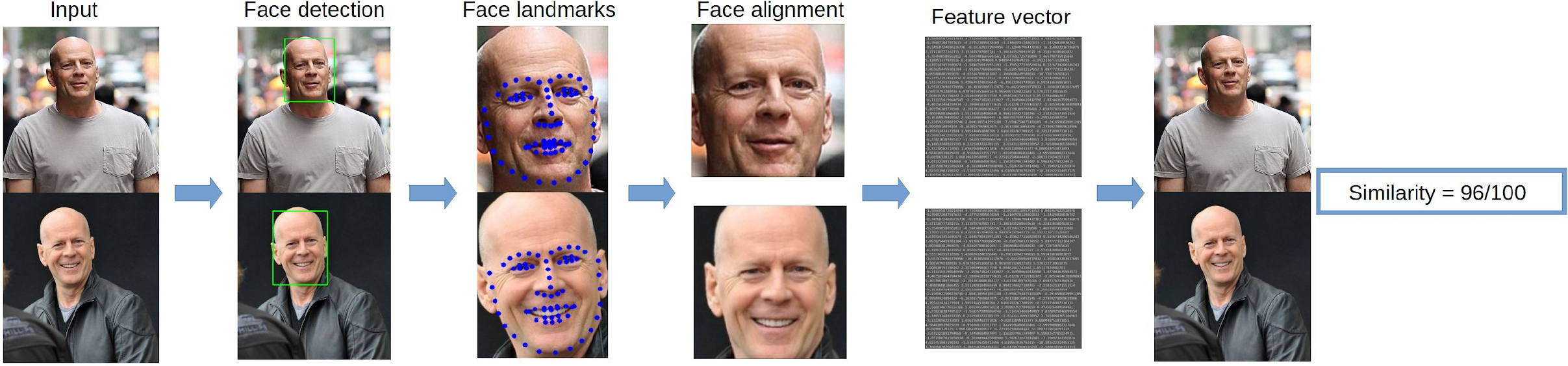}
 	\caption{Face recognition pipeline: ROI detection, landmark extraction, alignment, and feature matching.}
 	\label{fig:face_pipeline}
 \end{figure}

Face recognition pipelines usually require three components illustrated in Fig. \ref{fig:face_pipeline}: detection, alignment and feature vector extraction. The detector provides a bounding box for the face in a given image. For real-time face detection, we use a cascade type neural network based on \textit{MTCNN} \cite{Zhang2016}. The purpose of alignment is to normalize the face for the feature vector extractor to give more accurate and stable results on images taken in different conditions. To align a face to pre-defined coordinates, a face landmark detector can be used to find key points. The face is then transformed using a similarity transform (rotation, translation and scaling) to the target coordinates. For a fast face landmark detector, we use the \textit{LBF} \cite{Ren2016}. The aligned face image is then fed to a feature extractor neural network designed for recognition. For fast inference, we use a small and accurate neural network, based on \textit{MobileFaceNet} \cite{Chen2018}. This produces a low dimensional feature vector, unique for each identity. This feature vector is then compared to other vectors using cosine similarity to find matches.

\subsection{Multiple Person Tracking}
\label{subsec:mot}
Multi-object tracking aims at finding the trajectories of moving objects in a video sequence. It is crucial for video surveillance systems with obstructing objects and whose complexity increases with the number of humans to be tracked.  This problem is often treated as an association task. First, it marks out a rectangle that encloses each of the objects to be tracked (bounding box), and then a matching algorithm is responsible for associating each of the corresponding bounding boxes across the different frames.

In the last few years, the focus has been on tracking by detection in each of the frames. However, this method is computationally expensive because of the detection and does not usually meet the requirement of real-time performance  \cite{Hao2018}. Thus, in order to speed up the task, the detection is usually executed every few frames, and in between, another lighter algorithm is used. These widely used lighter algorithms include  traditional methods like mean shift, or Kalman filters which follow the location of the object through multiple consecutive iterations. However, after a number of iterations, these algorithms can result in a significant accumulated error, so they must be updated with the detector regularly.

In this work, tracking uses the movement and appearance of detected targets as in \cite{Wojke2017}. We run the person detection every 5 frames to update the tracking and meanwhile we match the identity of the different bounding boxes obtained by the background subtraction by predicting the location and the similarity in their appearance. For the prediction of trajectories within the frame and also across cameras we use Kalman filters. In contrast, for matching by appearance, a CNN obtains robust descriptors of the detected targets, that complements the biometric identification via facial recognition in case of e.g. ambiguity, or when a known target re-entries the scene.

\section{A reconfigurable CPS}
\label{sec:rCPS}
Reconfigurable CPSs provide solutions that automatically adapt their functionality or architectures to the dynamic environment. The purpose is to accomplish their tasks but also attains the optimization of specific qualities meanwhile. Quality and Resource Management (QRM) modules are in charge of monitoring the evolution of specific qualities or the environment and trigger reconfigurations accordingly.

\begin{table}[ht]
\centering
\caption{Edge operating modes}
\label{tab:edge_modes}
\begin{tabular}{@{}llll@{}}
\toprule
\multicolumn{1}{c}{Mode}      & \multicolumn{1}{c}{Resolution (fps)} & \multicolumn{1}{c}{Livestream video bandwidth} & \multicolumn{1}{c}{Event(s)}               \\ \midrule
{\color[HTML]{000000} }       & {\color[HTML]{000000} }        & {\color[HTML]{000000} } & {\color[HTML]{000000} Confirmed intrusion} \\
\multirow{-2}{*}{{\color[HTML]{000000} Mode 2}} &
  \multirow{-2}{*}{{\color[HTML]{000000} 1280x960 (30)}} &
  \multirow{-2}{*}{{\color[HTML]{000000} 8,40 MB/s}} &
  Broken perimeter \\
{\color[HTML]{000000} }       & {\color[HTML]{000000} }        &                         & Detection                                  \\
\multirow{-2}{*}{{\color[HTML]{000000} Mode 1}} &
  \multirow{-2}{*}{{\color[HTML]{000000} 640x480 (15)}} &
  \multirow{-2}{*}{1,90 MB/s} &
  Posible intrusion \\
{\color[HTML]{000000} Mode 0} & {\color[HTML]{000000} 320x240 (5)} & 0,41 MB/s                       & Nothing relevant occurs                             \\ \bottomrule
\end{tabular}%

\end{table}

Table \ref{tab:edge_modes} shows the reconfiguration scenarios with different modes that are analyzed in Section \ref{subsec:decisionmaking}. The QRM module drives reconfiguration monitoring events that happen in the scene and affects the resolution of the transmitted video, optimizing the use of bandwidth of the network that is shared by all the distributed local edges. The cloud-edge interaction for reconfiguration is done via reconfiguration commands from the cloud. Reconfiguration commands are responsible for establishing local edge modes of operation at every moment, modifying the quality of the video acquisition that is processed in the node and transmitted to the cloud. These reconfigurations have different direct effects: 1) image processing directly depends on image resolution; 2) the use of data bandwidth is crucial for a network shared by multiple cameras to ensure bounded latency and thus, our reconfigurable CPS reduces it significantly when nothing or only negligible events occur in the scene. It should be noted that the same video codec (MJPEG) has been used for all sequences.  All the results presented in this work, as well as the bandwidth values shown, have been calculated from the MJPEG-encoded video.

\subsection{Node Calibration}
\label{subsec:nodecalibration}

\begin{figure}[ht]
\centering
\includegraphics[scale=.50]{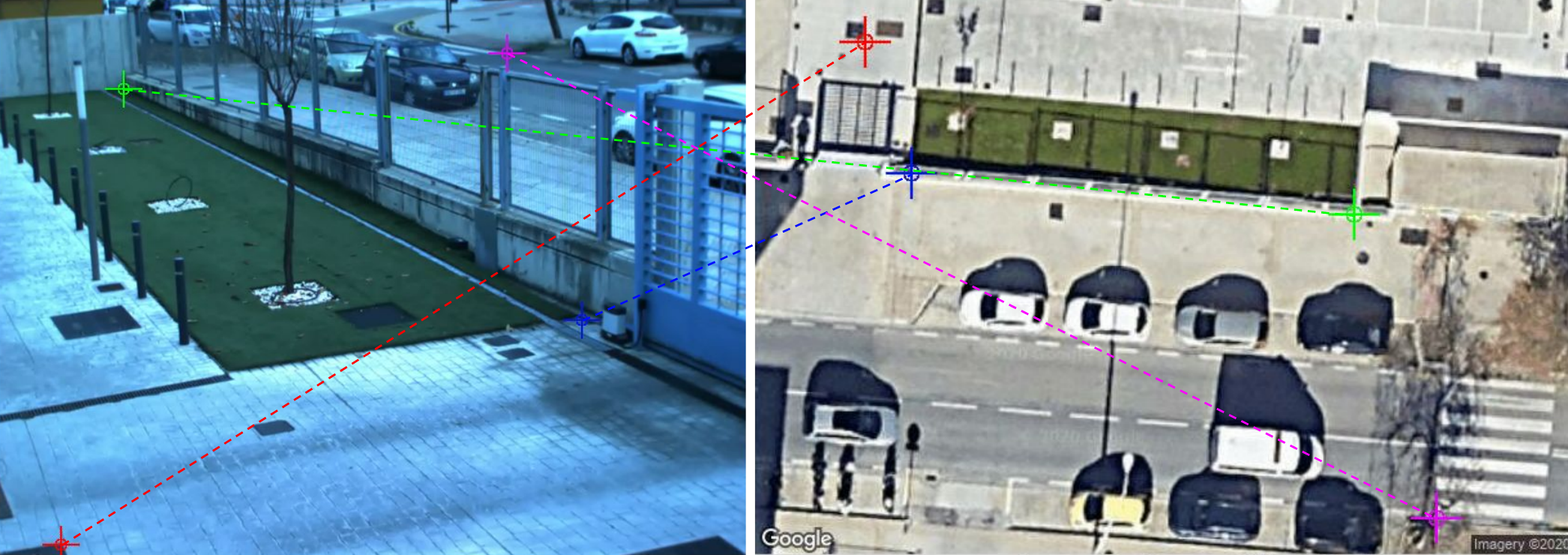}
\caption{The calibration tool computes the camera homographic parameters using the correspondences between the selected points from two different camera perspectives. In this example, only 4 points from two different views are required to compute the homography matrix (colors denote matches of points from the two views).}
\label{fig:calibration}
\end{figure}

As mentioned earlier, homography is the technique we use to calibrate cameras. For this purpose, points of the image and their equivalents on the ground plane have been selected by using cartographic coordinates. Fig. \ref{fig:calibration} shows an example of use with two cameras and point correspondences. We have released the calibration tool as open-source to help others in the field\footnote{https://github.com/JuanIsernGhosn/homography-calibrator}.

\subsection{Multi-camera decision making}
\label{subsec:decisionmaking}

\begin{figure}[ht]
\centering
\includegraphics[scale=.40]{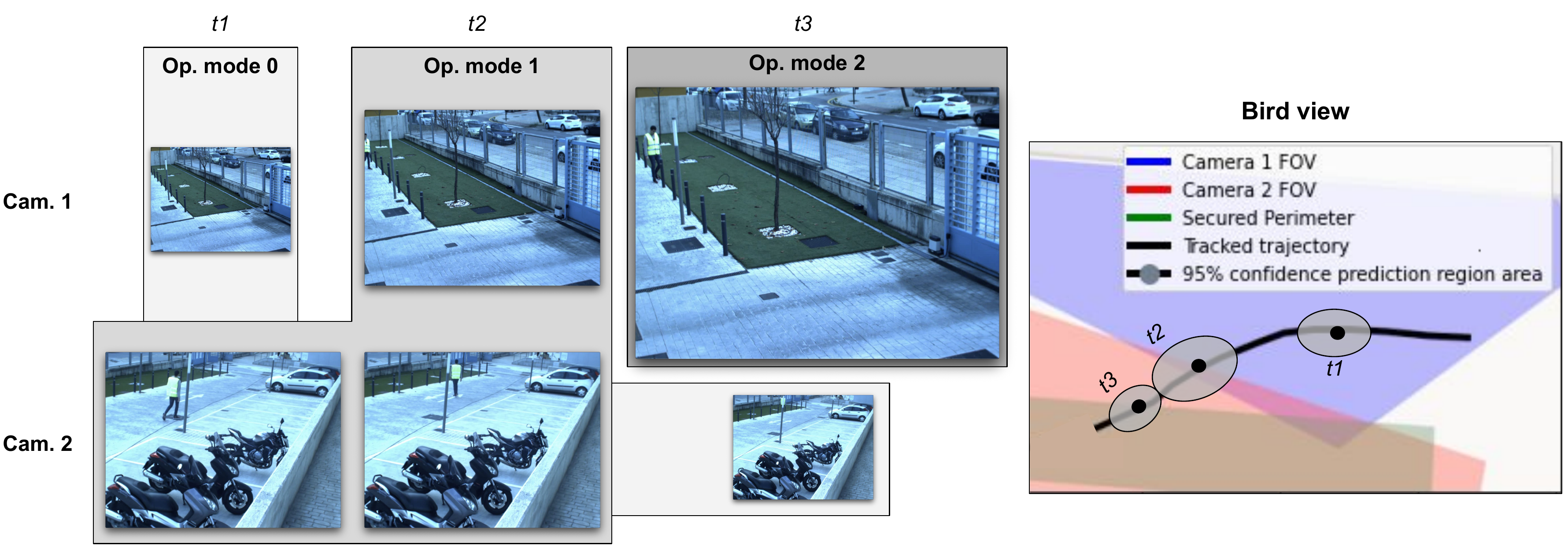}
\caption{Reconfiguration example for the multi-camera video-surveillance system: if nothing relevant happens (\textit{t1} in \textit{cam1} and \textit{t3} in \textit{cam2}), the node operates in mode 0 (see Table \ref{tab:edge_modes}); if a person is detected (detection or prediction of the trajectory within the FOV of at least one camera), mode 1 is activated (\textit{t2} in \textit{cam1} - blue area on the map - and \textit{t1} and \textit{t2} in \textit{cam2} - red area on the map); finally, mode 2 is activated with the intrusion in the secured perimeter (\textit{t3} in \textit{cam1} - green area on the map).}
\label{fig:decision_multi_camera}
\end{figure}

The reconfiguration is driven by the events detected in the scene. The multi-camera decision-making module is in charge of gathering the event data from the local edge nodes and trigger reconfiguration if needed. Data from the nodes include tracking (trajectories and target positions), people identification provided by facial recognition and deep robust features, and perimeter monitoring. As shown in the Table \ref{tab:edge_modes}, in the absence of events of interest in the FOV of a node resolution is reduced in order to minimize the amount of data to be processed and transferred. In the event of authorized personnel detected in the scene, they are supervised in the mid-priority operating mode (mode 1), while if there is an intrusion in the secured perimeter, the node(s) covering that region operates at maximum resolution to ensure the best accuracy in detection and identification (see Fig. \ref{fig:decision_multi_camera}). High resolution images are transmitted with a two-fold purpose: 1) allow the security personnel double-check potential risks, and 2) to save maximum resolution images for future forensic analysis.

The aspects considered to carry out the reconfiguration of the nodes are the following: When a person is detected within the node FOV (\textit{t1} person within the FOV of the camera of local node 1 (\textit{cam1}); blue area in Fig. \ref{fig:decision_multi_camera}-bottom), it changes its operation mode from 0 to 1. Next, when the trajectory of a tracked target points to the area of the predicted region with 95\% confidence (Ellipses show the potential region in which with 95\% confidence the track will be found in the next 200ms) being within the FOV of another camera (\textit{t2} with predicted trajectory within \textit{cam2} FOV; red area), the operation mode goes up to 2. Similarly, when the trajectory of a tracked target entries the perimeter of a secured area, the operation mode changes to 2 (\textit{t3}; green area).

\section{Implementation and results}
\label{sec:exp}

Local edge nodes are in charge of human detection within their FOV and the extraction of deep robust features from them. The cloud gathers information from all nodes, maintains the relationship between camera views, monitors secured perimeter, carries out tracking, and monitors the scene for triggering potential edge reconfigurations. Next subsections present the experiments and discuss results.    

Regarding the embedded devices for our edge nodes, we use a heterogeneous approach with Nvidia Jetson TX2 and Nvidia Jetson Xavier SoCs, bearing in mind their high performance and size. The Jetson TX2 module is a SoC with a six-core CPU with 8GB DDR4 memory and a GPU with NVidia Pascal architecture with 256 CUDA cores, while the Jetson Xavier is a Tegra Soc, with an 8-core CPU with 16GB DDR4 memory and a Volta GPU with 512 CUDA cores. Regarding the cloud, we are using a platform that includes a high-performance GPU NVidia RTX 2080Ti. 

\begin{table}[ht]
\centering
\caption{Performance on the edge for the different operating modes (fps).}
\label{tab:perf_edge}
\begin{tabular}{@{}lccc@{}}
\toprule
                & Mode 0  & Mode 1  & Mode 2 \\ \midrule
Jetson TX2      & 69.3    & 44.2    & 31.7 \\
Jetson Xavier   & 98.3    & 70.5    & 50.4 \\ \bottomrule
\end{tabular}%
\end{table}

To verify the compliance of the requirement of real-time operation, the performance of those tasks embedded on the edges is included in Table \ref{tab:perf_edge}. The performance in fps for each of the different operating modes and for the SoC devices considered reach in the worst case about 32 fps (for 1280x960 resolution images), and up to 70 fps in the best case at the lowest resolution. The Nvidia Jetson Xavier reaches 1.4x performance improve with respect to Jetson TX2. Processing is faster than video acquisition for all operating modes, allowing each edge node to connect to more than one camera at a time if needed.

\begin{figure}[t!]
\centering
\includegraphics[scale=.55]{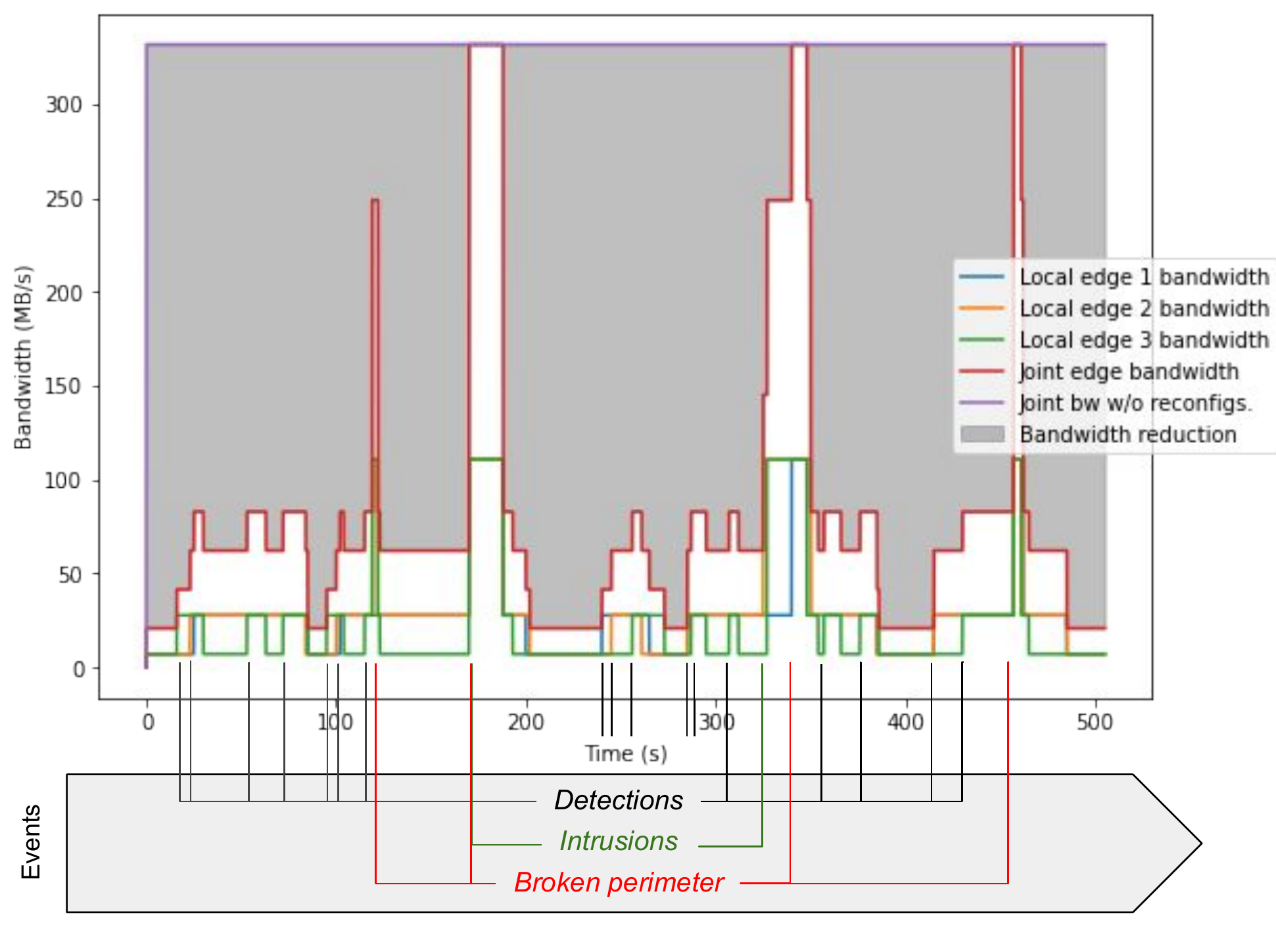}
\caption{Reconfiguration example for CPS cloud-edge bandwidth use. Three local edges are dynamically reconfigured according to the events (arrow) of the monitored environment. The gray area represents the bandwidth reduction ($\approx 75\%$) with respect to no reconfiguration scenario.}
\label{fig:banred}
\end{figure}

With respect to the overall system, reconfiguration is triggered to optimize the use of data bandwidth of the shared communication network in our multi-camera system. The use of data bandwidth directly depends on image resolution that is context-aware as shown in Section \ref{sec:rCPS}. Local nodes use the lowest resolution when no event happens, reducing communication bandwidth and thus, processing complexity. In contrast, when relevant events take place the resolution is improved to increase the detection accuracy and resolution of the images that are transferred to the cloud, changing the operation mode. Fig. \ref{fig:banred} shows a real example of how our system bandwidth changes and how the network load is reduced due to the reconfiguration of each one of the local edges. Different events (marked out in time axis) change the bandwidth use: if a node detects a human it is reconfigured and the bandwidth usage increases from 6.9 to 27.6 MB/s per node; if the human target is identified as non-authorized personnel or enters into a secured perimeter, it is again reconfigured and the bandwidth usage goes up to 110.6MB/s per node. In this example, reconfiguration allows reducing bandwidth by 76.32\%, reducing the video resolution mainly when no event of interest takes place in the monitored environment.

\subsection{Human detection}
\label{subsec:edge}
A comparison of two human detection models trained with Transfer Learning (\textit{MNV2\_1} and \textit{MNV2\_4}), the original \textit{MobileNetV2} model, and a \textit{HOG} (Histograms Of Gradients)\textit{+SVM} classic approach is shown in Table \ref{tab:classification}. The evaluation includes the INRIA Person Dataset and the OTC (Oxford Town Centre) video surveillance dataset. While MNV2\_1 uses the original architecture of MobileNetV2, adding at the end three dense layers with regularization, MNV2\_4 uses only the first three blocks of MobileNetV2, adding at the end only two dense layers with regularization. \textit{MNV2\_4} offers a good accuracy vs. resources trade-off, reaching almost the same accuracy than our first model \textit{MNV2\_1} and 17\% more than the original \textit{MobileNetV2} model and 3\% more than the classic \textit{HOG+SVM}. At the same time, it achieves 16x reduction compared to \textit{MobileNetV2} and 36x reduction to \textit{MNV2\_1} in the size of the model. This is a crucial parameter to implement the architecture in embedded devices, reducing complexity, processing time and power consumption. Also, to reduce even further the size and complexity of our CNN models, they have been optimized through TensorRT.

\begin{table}[ht]
\centering
\caption{Human detection performance (INRIA + OTC datasets)}
\label{tab:classification}
\begin{tabular}{@{}llll@{}}
\toprule
\multicolumn{1}{c}{Model name}          & \multicolumn{1}{c}{F1-Score}             & Neural network parameters                   & Disk space (MB)                    \\ \midrule
{\color[HTML]{000000} MobileNetV2}          & {\color[HTML]{000000} 0.801574} & {\color[HTML]{000000} 3.4 M} & {\color[HTML]{000000} 14.0} \\
{\color[HTML]{000000} MNV2\_1}          & {\color[HTML]{000000} \textbf{0.970489}} & {\color[HTML]{000000} 4.1 M} & {\color[HTML]{000000} 31.60} \\
{\color[HTML]{000000} \textbf{MNV2\_4}} & {\color[HTML]{000000} 0.964618}          & \textbf{75,186}              & \textbf{0.88}                \\
{\color[HTML]{000000} HOG + SVM} & {\color[HTML]{000000} 0.947584}          & -              & 140.4                \\
 \bottomrule
\end{tabular}%
\end{table}

\subsection{Facial recognition}
\label{subsec:facialperf}

 \begin{figure}[ht]
 	\centering
 	\includegraphics[scale=.45]{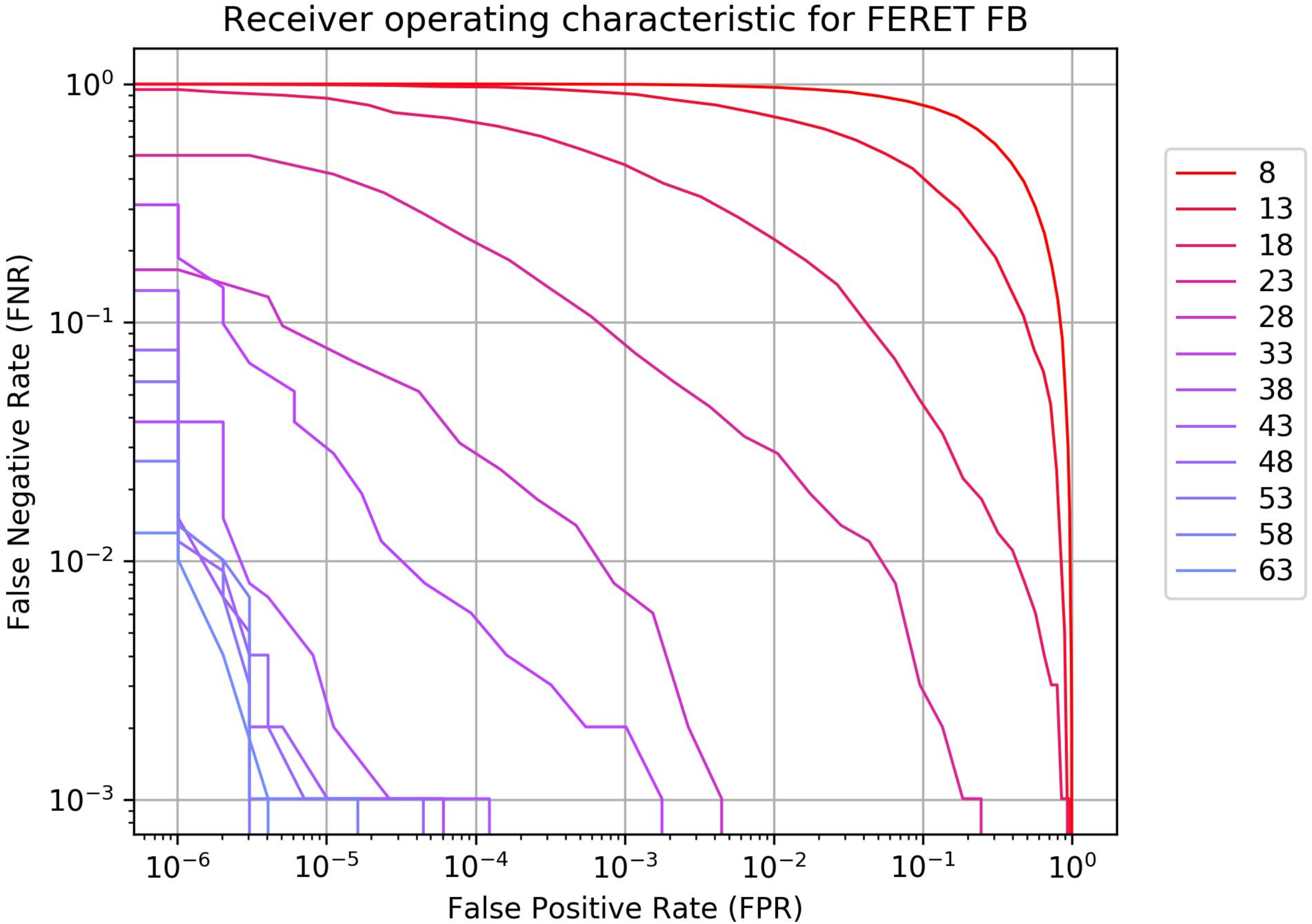}
 	\caption{False positive rate as a function of false negative rate of face recognition for the FERET FB test set \cite{Phillips2000}, with varying resolution of faces. The figure represents the plots with sizes of 8-63 pixels per box side.
 	}
 	\label{fig:roc_vs_res_face_rec}
 \end{figure}

Our implementation of \textit{MTCNN} face detector was trained using two datasets \cite{Yang2016, Sun2013} following the original \textit{MTCNN} training. The face landmark detector (LFB), was trained using the Helen \cite{Le2012}, the LFPW \cite{BelHumeur2013} and the iBug \cite{Sagonas2013} datasets. The feature extractor \textit{MobileFaceNet} was trained on the \textit{MS-Celeb-1M} dataset \cite{Yandong2016}. The accuracy of our face recognition pipeline on the LFW test set is $99.50\%$. On a modern smartphone (One+ 7T Pro) the inference speed is $\approx 30ms$ per frame. The resolution of a face image in a face recognition pipeline is critical for the accuracy as shown in Figure \ref{fig:roc_vs_res_face_rec}. The smallerhttps://www.overleaf.com/project/5fbe7fb61a454806e05bec0c the face, the less accurate will the face detection and alignment be, resulting in less accurate face recognition overall. Thus, when the size of the face images is 8 pixels, the FNR(@FPR$<$$10^{-6}$)=100.0\%, whereas when they are 63 pixels long, the FNR(@FPR$<$$10^{-6}$)=0.001\%, which is $~10^5$ times better than using the lower resolution images.

\subsection{Tracking}
\label{subsec:trackingperf}

In order to evaluate the quality of the tracking, we present a comparison with the state-of-the-art FairMOT method \cite{zhang2020simple}. It achieves the best results in the MOT challenge for the MOT20 dataset \cite{MOTChallenge20}. The comparison is shown in Table \ref{tab:one_cam_mot} runs on the VIRAT dataset \cite{Oh2011}, with surveillance sequences in single-camera environments. The evaluation is carried out using the metrics: 1) Multi-object tracking accuracy (MOTA), overall tracking accuracy in terms of false positives, false negatives and identity switches; 2) Multi-object tracking precision (MOTP), overall tracking precision in terms of bounding box overlap between ground-truth and reported location; 3) Mostly tracked (MT), percentage of ground-truth tracks that have the same label for at least 80\% of their life span; 4) Mostly lost(ML), percentage of ground-truth tracks that are tracked for at most 20\% of their life span; 5) Identity switches (ID), number of times the reported identity of a ground-truth track changes.

\begin{table}[ht]
\centering
\caption{Multiple Object Tracking (MOT) results in sequences of the VIRAT dataset.}
\label{tab:one_cam_mot}
\begin{tabular}{@{}lccccccc@{}}
\toprule
 &
  {\color[HTML]{000000} \textbf{MOTA $\uparrow$}} &
  {\color[HTML]{000000} \textbf{MOTP $\uparrow$}} &
  \textbf{MT $\uparrow$} &
  \textbf{ML $\downarrow$} &
  \textbf{FN $\downarrow$} &
  \textbf{FP $\downarrow$} &
  \textbf{ID $\downarrow$}\\ \midrule
Ours &
  {\color[HTML]{000000} 89.4} &
  92.8 &
  {\color[HTML]{000000} 90.3\%} &
  4.3\% &
  {\color[HTML]{000000} 5.1\%} &
  3.8\% &
  1.8\% \\
FairMOT \cite{zhang2020simple} &
  \textbf{94.2} &
  \textbf{96.9} &
  \textbf{93.7\%} &
  \textbf{2.5\%} &
  \textbf{3.3\%} &
  \textbf{1.9\%} &
  \textbf{0.6\%} \\
 DeepMOT \cite{xu2019train} &
  91.4 &
  95.8 &
  92.8 &
  2.6\% &
  3.4\% &
  2.4\% &
  \textbf{0.6\%} \\ \bottomrule
\end{tabular}%
\end{table}

Table \ref{tab:one_cam_mot} shows the tracking results of our system with respect to two of the best trackers in the state of the art. All these results are obtained in a single-camera environment. In general terms, the result of our system regarding these two pioneer trackers is good. The number of false negatives is small, an important feature of  video surveillance systems whose main purpose is to minimize triggering false alarms. Moreover, compared to the other trackers, our system guarantees its operation in real time reaching up to 50.4 fps in HQ video (compared to the others that reach 30 fps but using a dedicated powerful GPU). Since FairMOT is the state of the art method tracking in single-camera environments, it has been used to build our groundtruth, using its results to label the videos in our own test dataset. However, since our dataset corresponds to video sequences from a multi-camera environment, a manual matching of the track IDs between different perspectives was also required.

The results for multiple object tracking in our multi-camera test dataset are shown in Table \ref{tab:multi_cam_mot}. The main issue is the complete lack of appropriate datasets and methods for multi-camera setups. In order to compare our multi-camera system we gathered the results of multiple object tracking obtained by DeepMot in each video sequence from each camera in our dataset, as if they were multiple videos from a single camera dataset. Then, we averaged the resulting MOT metrics for each video sequence. The difference between results is not significant for both systems (MOTA: 1\%, FN: 0.5\%, FP: 0.2\%), with a  slight advantage of DeepMot. However, since part of the processing is distributed in the local nodes at the edge, our system continues working in real time even with the processing of multiple video streams at the same time and maintaining the same performance.

\begin{table}[ht]
\centering
\caption{MOT results for our test dataset}
\label{tab:multi_cam_mot}
\begin{threeparttable}
\begin{tabular}{@{}lccccccc@{}}
\toprule
 &
  {\color[HTML]{000000} \textbf{MOTA $\uparrow$}} &
  {\color[HTML]{000000} \textbf{MOTP $\uparrow$}} &
  \textbf{MT $\uparrow$} &
  \textbf{ML $\downarrow$} &
  \textbf{FN $\downarrow$} &
  \textbf{FP $\downarrow$} &
  \textbf{ID $\downarrow$} \\ \midrule
Ours* &
  {\color[HTML]{000000} 98.4} &
  97.2 &
  {\color[HTML]{000000} 96.0\%} &
  2.2\% &
  {\color[HTML]{000000} 0.9\%} &
  0.5\% &
  0.1\% \\ 
DeepMOT** &
  99.4 &
  98.5 &
  98.7\% &
  0.2\% &
  0.4\% &
  0.3\% &
  0.1\% \\ \bottomrule
\end{tabular}
\begin{tablenotes}\footnotesize
\item[*] Result considering the multi-camera setup and the same detection identities across different perspectives for each video sequence.
\item[**] It's a single-camera tracker. Average of the results in each perspective of the sequences of our multi-camera dataset
\end{tablenotes}
\end{threeparttable}
\end{table}

\section{Conclusions}
Our work has novel differentiating aspects that are necessary in modern video-surveillance systems for Smart Cities. First, our active reconfigurable CPS reacts automatically to the dynamic events that occur in the monitored environment. The real time requirements demands an appropriate management of the resources, without harming the accuracy of the system. In our case, the intelligent management of the bandwidth shared by the system nodes, allows to reduce the average bandwidth usage in approximately 75\%. The resolution reduction does not affect biometric identification and high resolution images are only transmitted to reduce false positive detections and for forensic purposes.

Second, machine learning modules have been optimized and embedded in SoCs for the detection and characterization of people within the monitored area. Using Deep Learning techniques, we have proposed new CNN-based models using Transfer Learning for the video-surveillance field that lacks appropriate datasets. These models have been optimized using TensorRT, achieving real-time performance. We process high quality video streams above 30~fps on Nvidia Jetson TX2 and above 50~fps on Nvidia Jetson Xavier platforms. This facilitates integrating multiple cameras into single nodes working at lower framerates, if required.

Finally, the efficient management of processing and communication resources using a reconfiguration scheme driven by a quality management system facilitates the scalability of the proposed approach. In addition, our system enables easy calibration of local edge nodes. By calibrating on a common coordinate system, it is possible to execute in a more robust way tasks such as multiple person tracking, allowing to exchange information from different local nodes to improve its performance.

\section*{Acknowledgments}
This work was partially supported by the EU Project FitOptiVis through the ECSEL Joint Undertaking under GA n. 783162, a Spanish National grant funded by MINECO through APCIN PCI2018‐093184, and partially by the Research Network RED2018-102511-T, the AEI Grant PID2019-109434RA-I00, and the Program 9 of the 2015 research plan of the University of Granada (project ID 16).

\bibliographystyle{unsrt}  
\bibliography{references} 

\begin{thebibliography}{10}

\bibitem{washburn2009helping}
Doug Washburn, Usman Sindhu, Stephanie Balaouras, Rachel~A Dines, N~Hayes, and
  Lauren~E Nelson.
\newblock Helping cios understand “smart city” initiatives.
\newblock {\em Growth}, 17(2):1--17, 2009.

\bibitem{Gupta2014}
Raghav Gupta, Ratish Agarwal, and Sachin Goyal.
\newblock {A Review of Cyber Security Techniques for Critical Infrastructure
  Protection}.
\newblock {\em International Journal of Computer Science {\&} engineering
  Technology (IJCTSET)}, 2014.

\bibitem{Talari2017}
Saber Talari, Miadreza Shafie-Khah, Pierluigi Siano, Vincenzo Loia, Aurelio
  Tommasetti, and Jo{\~{a}}o~P.S. Catal{\~{a}}o.
\newblock {A review of smart cities based on the internet of things concept}.
\newblock {\em Energies}, 10(4):1--23, 2017.

\bibitem{Cenedese2014}
Angelo Cenedese, Andrea Zanella, Lorenzo Vangelista, and Michele Zorzi.
\newblock {Padova Smart City: An urban Internet of Things experimentation}.
\newblock In {\em Proceeding of IEEE International Symposium on a World of
  Wireless, Mobile and Multimedia Networks 2014}, pages 1--6. IEEE, jun 2014.

\bibitem{Krizhevsky2012}
A.~{Krizhevsky}, I.~{Sutskever}, and G.~{Hinton}.
\newblock Imagenet classification with deep convolutional neural networks.
\newblock In F.~Pereira, C.~J.~C. Burges, L.~Bottou, and K.~Q. Weinberger,
  editors, {\em Advances in Neural Information Processing Systems 25}, pages
  1097--1105. Curran Associates, Inc., 2012.

\bibitem{Kalchbrenner2014}
Nal Kalchbrenner, Edward Grefenstette, and Phil Blunsom.
\newblock {A Convolutional Neural Network for Modelling Sentences}.
\newblock In {\em Proceedings of the 52nd Annual Meeting of the Association for
  Computational Linguistics (Volume 1: Long Papers)}, pages 655--665,
  Stroudsburg, PA, USA, 2014. Association for Computational Linguistics.

\bibitem{Barra2020}
Silvio Barra, Salvatore~Mario Carta, Andrea Corriga, Alessandro~Sebastian
  Podda, and Diego~Reforgiato Recupero.
\newblock {Deep learning and time series-To-image encoding for financial
  forecasting}.
\newblock {\em IEEE/CAA Journal of Automatica Sinica}, 2020.

\bibitem{Liu2019}
Yongcheng Liu, Bin Fan, Shiming Xiang, and Chunhong Pan.
\newblock {Relation-Shape Convolutional Neural Network for Point Cloud
  Analysis}.
\newblock In {\em 2019 IEEE/CVF Conference on Computer Vision and Pattern
  Recognition (CVPR)}, pages 8887--8896. IEEE, jun 2019.

\bibitem{Abiwinanda2019}
Nyoman Abiwinanda, Muhammad Hanif, S.~Tafwida Hesaputra, Astri Handayani, and
  Tati~Rajab Mengko.
\newblock {Brain Tumor Classification Using Convolutional Neural Network}.
\newblock In {\em IFMBE Proceedings}, pages 183--189. 2019.

\bibitem{LFW2007}
Gary~B. Huang, Manu Ramesh, Tamara Berg, and Erik Learned-Miller.
\newblock Labeled faces in the wild: A database for studying face recognition
  in unconstrained environments.
\newblock Technical Report 07-49, University of Massachusetts, Amherst, October
  2007.

\bibitem{Turk1991}
M.~{Turk} and A.~{Pentland}.
\newblock Eigenfaces for recognition.
\newblock {\em Journal of Cognitive Neuroscience}, 3(1):71--86, 1991.

\bibitem{Ahonen2006}
T.~{Ahonen}, A.~{Hadid}, and M.~{Pietikainen}.
\newblock Face description with local binary patterns: Application to face
  recognition.
\newblock {\em IEEE Transactions on Pattern Analysis and Machine Intelligence},
  28(12):2037--2041, Dec 2006.

\bibitem{Abate2020}
Andrea~F. Abate, Paola Barra, Silvio Barra, Cristiano Molinari, Michele Nappi,
  and Fabio Narducci.
\newblock {Clustering Facial Attributes: Narrowing the Path From Soft to Hard
  Biometrics}.
\newblock {\em IEEE Access}, 8:9037--9045, 2020.

\bibitem{liu2015faceattributes}
Ziwei Liu, Ping Luo, Xiaogang Wang, and Xiaoou Tang.
\newblock Deep learning face attributes in the wild.
\newblock In {\em Proceedings of International Conference on Computer Vision
  (ICCV)}, December 2015.

\bibitem{Haering2008}
Niels Haering, P{\'{e}}ter~L. Venetianer, and Alan Lipton.
\newblock {The evolution of video surveillance: An overview}.
\newblock {\em Machine Vision and Applications}, 2008.

\bibitem{Chen2018}
Ning Chen and Yu~Chen.
\newblock {Smart City Surveillance at the Network Edge in the Era of IoT:
  Opportunities and Challenges}.
\newblock In {\em Smart Cities}, pages 153--176. 2018.

\bibitem{Masin2017}
M~Masin, F~Palumbo, H~Myrhaug, J.~A. {de Oliveira Filho}, M~Pastena, M~Pelcat,
  L~Raffo, F~Regazzoni, A~A Sanchez, A~Toffetti, E.~de~la Torre, and K~Zedda.
\newblock {Cross-layer design of reconfigurable cyber-physical systems}.
\newblock In {\em DATE}, pages 740--745. IEEE, mar 2017.

\bibitem{Salman2018}
Bakhita Salman, Mohammed~I. Thanoon, Saleh Zein-Sabatto, and Fenghui Yao.
\newblock {Multi-camera Smart Surveillance System}.
\newblock {\em Int. Conf. on Computational Science and Computational
  Intelligence}, pages 468--472, 2018.

\bibitem{Scionti2019}
A.~Scionti, S.~Ciccia, O.~Terzo, and G.~Giordanengo.
\newblock {Chip-to-Cloud: An Autonomous and Energy Efficient Platform for Smart
  Vision Applications}.
\newblock In {\em Design, Automation and Test in Europe Conference}, pages
  492--497, 2019.

\bibitem{Piccardi2004}
Massimo Piccardi.
\newblock {Background subtraction techniques: a review}.
\newblock In {\em IEEE Int. Conf. on Systems, Man and Cybernetics}, volume~4,
  pages 3099--3104. IEEE, 2004.

\bibitem{Zivkovic2004}
Zoran Zivkovic.
\newblock {Improved adaptive Gaussian mixture model for background
  subtraction}.
\newblock In {\em International Conference on Pattern Recognition}, volume~2,
  pages 28--31, 2004.

\bibitem{Park2020}
Jisoo Park, Jingdao Chen, Yong~K. Cho, Dae~Y. Kang, and Byung~J. Son.
\newblock {CNN-based person detection using infrared images for night-time
  intrusion warning systems}.
\newblock {\em Sensors (Switzerland)}, 20(1), 2020.

\bibitem{Zhao2019}
Congcong Zhao and Bin Chen.
\newblock {Real-Time Pedestrian Detection Based on Improved YOLO Model}.
\newblock {\em Conf. on Human-Machine Systems and Cybernetics}, 2:25--28, 2019.

\bibitem{imagenet_cvpr09}
J.~Deng, W.~Dong, R.~Socher, L.-J. Li, K.~Li, and L.~Fei-Fei.
\newblock {ImageNet: A Large-Scale Hierarchical Image Database}.
\newblock In {\em CVPR09}, 2009.

\bibitem{Sandler2018}
Mark Sandler, Andrew Howard, Menglong Zhu, Andrey Zhmoginov, and Liang~Chieh
  Chen.
\newblock {MobileNetV2: Inverted Residuals and Linear Bottlenecks}.
\newblock {\em Computer Vision and Pattern Recognition}, pages 4510--4520,
  2018.

\bibitem{Long2019}
Li~Long and Shan Dongri.
\newblock {Review of Camera Calibration Algorithms}.
\newblock pages 723--732. 2019.

\bibitem{Zhang2000}
Zhengyou Zhang.
\newblock {A flexible new technique for camera calibration}.
\newblock {\em IEEE Transactions on Pattern Analysis and Machine Intelligence},
  2000.

\bibitem{Zhang2016}
K.~{Zhang}, Z.~{Zhang}, Z.~{Li}, and Y.~{Qiao}.
\newblock Joint face detection and alignment using multitask cascaded
  convolutional networks.
\newblock {\em IEEE Signal Processing Letters}, 23(10):1499--1503, Oct 2016.

\bibitem{Ren2016}
Shaoqing Ren, Xudong Cao, Yichen Wei, and Jian Sun.
\newblock {Face Alignment via Regressing Local Binary Features}.
\newblock {\em IEEE Transactions on Image Processing}, 25(3):1233--1245, mar
  2016.

\bibitem{Hao2018}
Tong Hao, Qian Wang, Dan Wu, and Jin~Sheng Sun.
\newblock {Multiple person tracking based on slow feature analysis}.
\newblock {\em Multimedia Tools and Applications}, 77(3):3623--3637, 2018.

\bibitem{Wojke2017}
Nicolai Wojke, Alex Bewley, and Dietrich Paulus.
\newblock {Simple online and realtime tracking with a deep association metric}.
\newblock In {\em ICIP}, pages 3645--3649. IEEE, sep 2017.

\bibitem{Phillips2000}
P.~J. {Phillips}, {Hyeonjoon Moon}, S.~A. {Rizvi}, and P.~J. {Rauss}.
\newblock The feret evaluation methodology for face-recognition algorithms.
\newblock {\em IEEE Transactions on Pattern Analysis and Machine Intelligence},
  22(10):1090--1104, Oct 2000.

\bibitem{Yang2016}
S.~{Yang}, P.~{Luo}, C.~C. {Loy}, and X.~{Tang}.
\newblock Wider face: A face detection benchmark.
\newblock In {\em Conf. on Computer Vision and Pattern Recognition}, pages
  5525--5533, June 2016.

\bibitem{Sun2013}
Y.~{Sun}, X.~{Wang}, and X.~{Tang}.
\newblock Deep convolutional network cascade for facial point detection.
\newblock In {\em CVPR}, pages 3476--3483, June 2013.

\bibitem{Le2012}
Vuong Le, Jonathan Brandt, Zhe Lin, Lubomir Bourdev, and Thomas~S. Huang.
\newblock Interactive facial feature localization.
\newblock In Andrew Fitzgibbon, Svetlana Lazebnik, Pietro Perona, Yoichi Sato,
  and Cordelia Schmid, editors, {\em Computer Vision -- ECCV 2012}, pages
  679--692, Berlin, Heidelberg, 2012. Springer Berlin Heidelberg.

\bibitem{BelHumeur2013}
P.~N. {Belhumeur}, D.~W. {Jacobs}, D.~J. {Kriegman}, and N.~{Kumar}.
\newblock Localizing parts of faces using a consensus of exemplars.
\newblock {\em IEEE Transactions on Pattern Analysis and Machine Intelligence},
  35(12):2930--2940, Dec 2013.

\bibitem{Sagonas2013}
C.~{Sagonas}, G.~{Tzimiropoulos}, S.~{Zafeiriou}, and M.~{Pantic}.
\newblock 300 faces in-the-wild challenge: The first facial landmark
  localization challenge.
\newblock In {\em Int. Conference on Computer Vision Workshops}, pages
  397--403, Dec 2013.

\bibitem{Yandong2016}
Yandong Guo, Lei Zhang, Yuxiao Hu, Xiaodong He, and Jianfeng Gao.
\newblock Ms-celeb-1m: Challenge of recognizing one million celebrities in the
  real world.
\newblock {\em Electronic Imaging}, 2016:1--6, 02 2016.

\bibitem{zhang2020simple}
Yifu Zhang, Chunyu Wang, Xinggang Wang, Wenjun Zeng, and Wenyu Liu.
\newblock A simple baseline for multi-object tracking.
\newblock {\em arXiv preprint arXiv:2004.01888}, 2020.

\bibitem{MOTChallenge20}
P.~Dendorfer, H.~Rezatofighi, A.~Milan, J.~Shi, D.~Cremers, I.~Reid, S.~Roth,
  K.~Schindler, and L.~Leal-Taix\'{e}.
\newblock Mot20: A benchmark for multi object tracking in crowded scenes.
\newblock {\em arXiv:2003.09003[cs]}, March 2020.

\bibitem{Oh2011}
Sangmin Oh, Anthony Hoogs, Amitha Perera, Naresh Cuntoor, Chia-Chih Chen,
  Jong~Taek Lee, Saurajit Mukherjee, J.~K. Aggarwal, Hyungtae Lee, Larry Davis,
  Eran Swears, Xioyang Wang, Qiang Ji, Kishore Reddy, Mubarak Shah, Carl
  Vondrick, Hamed Pirsiavash, Deva Ramanan, Jenny Yuen, Antonio Torralba,
  Bi~Song, Anesco Fong, Amit Roy-Chowdhury, and Mita Desai.
\newblock {A large-scale benchmark dataset for event recognition in
  surveillance video}.
\newblock In {\em CVPR 2011}, pages 3153--3160. IEEE, jun 2011.

\bibitem{xu2019train}
Yihong Xu, Aljo\v{s}a O\v{s}ep, Yutong Ban, Radu Horaud, Laura Leal-Taix{\'e},
  and Xavier Alameda-Pineda.
\newblock How to train your deep multi-object tracker.
\newblock In {\em Conference on Computer Vision and Pattern Recognition
  (CVPR)}, 2020.

\end{thebibliography}
\end{document}